\documentclass{article} 
\usepackage{arxiv_preprint,times}

\usepackage{amsmath,amsfonts,bm}

\def\eqref#1{equation~\ref{#1}}

\def\1{\bm{1}}

\def\va{{\bm{a}}}

\def\ve{{\bm{e}}}

\def\vo{{\bm{o}}}
\def\vp{{\bm{p}}}
\def\vq{{\bm{q}}}

\def\vu{{\bm{u}}}

\def\vz{{\bm{z}}}

\def\mA{{\bm{A}}}

\def\mM{{\bm{M}}}

\def\mW{{\bm{W}}}

\def\mZ{{\bm{Z}}}

\DeclareMathAlphabet{\mathsfit}{\encodingdefault}{\sfdefault}{m}{sl}
\SetMathAlphabet{\mathsfit}{bold}{\encodingdefault}{\sfdefault}{bx}{n}

\def\sA{{\mathbb{A}}}

\newcommand{\R}{\mathbb{R}}

\usepackage{amsmath,amssymb,mathtools,bm}
\usepackage{xcolor}
\usepackage{amsthm}
\usepackage{graphicx}
\usepackage{booktabs}
\usepackage{multirow}
\usepackage{microtype}
\usepackage{hyperref}
\usepackage{url}
\usepackage{float}
\usepackage{placeins}
\usepackage{capt-of}

\newcommand{\fastlewm}{\textsc{Fast-LeWM}}

\newcommand{\cL}{\mathcal{L}}

\newcommand{\sg}{\operatorname{sg}}

\newtheorem{theorem}{Theorem}

\title{Parallel Predictive World Models for \\ Accurate and Efficient Long-Horizon Planning}

\author{
\begin{tabular}{c}
Wanjin Feng\textsuperscript{1} \quad
Baobin Zhang\textsuperscript{2,*} \quad
Ao Yu\textsuperscript{3} \quad
Shibo Feng\textsuperscript{4} \quad
Xi Wang\textsuperscript{2} \quad
Xingyu Gao\textsuperscript{2,*} \\
{\normalfont\small \textsuperscript{1}Tsinghua University} \\
{\normalfont\small \textsuperscript{2}Institute of Microelectronics, Chinese Academy of Sciences} \\
{\normalfont\small \textsuperscript{3}The Hong Kong University of Science and Technology (Guangzhou)} \\
{\normalfont\small \textsuperscript{4}Nanyang Technological University, Singapore} \\
{\normalfont\small \textsuperscript{*}Corresponding authors}
\end{tabular}
}

\begin{document}

\maketitle

\begin{abstract}
Long-horizon world-model planning typically relies on autoregressive rollouts, where predicted states are repeatedly fed back into the model.
This preserves temporal structure but creates a horizon-length sequential path and exposes later predictions to recursive decoded-state feedback.
We introduce Parallel Predictive World Models (PPWM), which predict a finite-horizon trajectory in parallel while retaining causal interaction among future representations.
Each horizon is conditioned on its causal action prefix, and future representations interact before decoding, separating temporal causality from state-by-state output recursion.
We formalize this distinction by viewing autoregressive rollout as a causal trajectory map and identifying the decoded-state feedback pathway removed by PPWM.
Across four visual-control tasks, PPWM achieves the lowest long-horizon prediction error and the highest Cross-Entropy Method (CEM) simulator success among the evaluated predictive interfaces.
Meanwhile, PPWM achieves more than a $3\times$ average CEM planning speedup over the autoregressive LeWM baseline.
These results suggest that accurate and efficient long-horizon world-model planning does not require state-by-state autoregression, but can instead be achieved through parallel causal trajectory prediction.
\end{abstract}

\section{Introduction}
\label{sec:intro}

World models let agents evaluate candidate behaviors through imagined futures before acting. Effective planning often requires sufficiently long futures, because the consequences of early actions may emerge only many steps later. Learned dynamics have therefore been paired with trajectory optimization and model-predictive control~\citep{chua2018pets,hansen2022tdmpc,hansen2024tdmpc2}, while recent sequence and visual world models extend this paradigm to high-dimensional observations and offline planning~\citep{micheli2023iris,zhou2025dinowm,sobal2025pldm}. As horizons grow, world-model planning must preserve predictive accuracy while avoiding the increasing sequential cost of state-by-state rollout.

Most world models expose future dynamics through recursive local prediction. Distant states require a horizon-dependent chain of model evaluations, and each predicted state becomes part of the next model input, allowing prediction error or model misspecification to propagate through the rollout~\citep{hafner2019planet,janner2019mbpo,gao2026fastlewm,somalwar2025imperfect}. Direct multi-step prediction provides an alternative factorization that can mitigate recursive bias under partial observability or model misspecification~\citep{somalwar2025imperfect,somalwar2026statistical}. Crucially, downstream planners inherit the same temporal factorization when evaluating long trajectories or propagating the influence of early actions to distant outcomes.

Recent work reduces this dependency from two sides. Prediction-side methods directly parameterize longer futures~\citep{mishra2017temporal}. Most closely, Fast-LeWM accumulates action effects with causal action prefixes and predicts multiple horizons in parallel~\citep{gao2026fastlewm}, but does not explicitly couple future-state representations before decoding. Planning-side methods instead lift intermediate states to expose trajectory-level parallelism~\citep{rybkin2021latco,ziakas2026gvpwm,wu2026koopmanboxqp}. GRASP performs parallel iterative optimization with virtual states while avoiding long chains of state-input Jacobians through an action-gradient formulation~\citep{psenka2026grasp}. 
This leaves a concrete design problem: preserve trajectory causality without forcing prediction and planning through a state-by-state temporal chain.

We introduce Parallel Predictive World Models (PPWM) for long-horizon planning.
Given recent context and an action span, PPWM predicts a causally structured finite-horizon trajectory directly, without output-state recursion inside the span and with causal interaction among future representations.
This design induces a parallel trajectory interface that can be queried by population-based search and differentiated jointly for lightweight trajectory-wide refinement.
In closed-loop control, the refined plan is repeatedly re-anchored by fresh observations rather than optimized once and executed open loop.
By removing recursive decoded-state feedback within each span, PPWM targets both long-horizon predictive accuracy and planning efficiency: it removes one recursive error-propagation path and shortens the sequential dependency exposed to planning. We evaluate these effects through prediction accuracy, simulator planning, closed-loop refinement, architectural generality, and wall-clock scaling.

\begin{figure}[t]
\centering
\includegraphics[width=\linewidth]{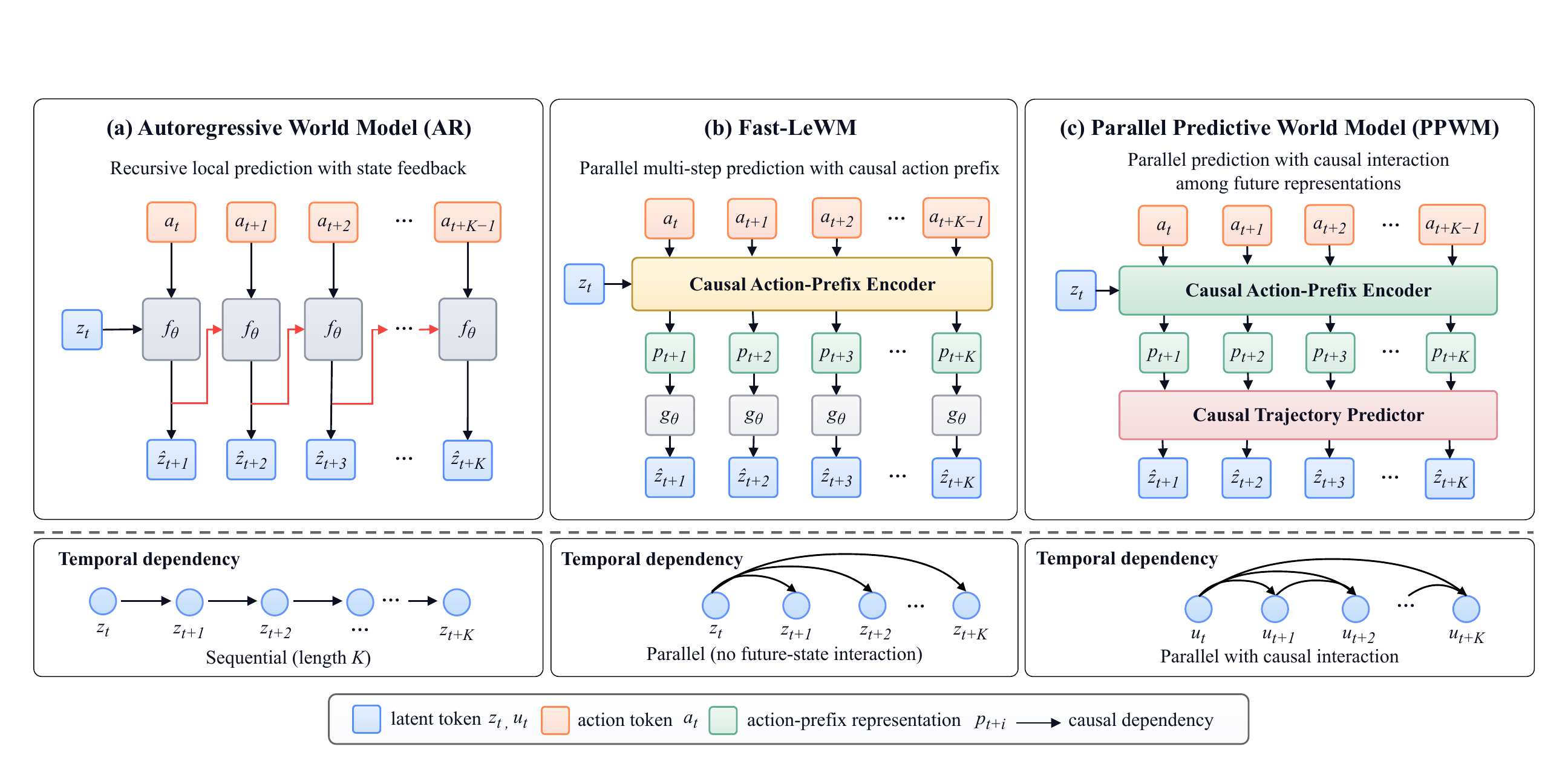}
\caption{Comparison of temporal interfaces for world models. (a) Autoregressive (AR) world models predict one step at a time and feed the predicted state back as input, forming a long temporal chain. (b) Fast-LeWM uses a causal action prefix to predict multiple future states in parallel, without explicit interaction among future-state representations. (c) Our Parallel Predictive World Model (PPWM) predicts a causal finite-horizon trajectory in parallel, allowing interaction among future representations while removing output-state recursion within a span.}
\label{fig:overview}
\end{figure}

Our contributions are:
\begin{itemize}
    \item \textbf{Parallel Predictive World Models.}
    We introduce PPWM, which replaces state-by-state decoded prediction with parallel causal trajectory prediction while preserving interaction among future representations.

    \item \textbf{Causal trajectory factorization for planning.}
    We characterize autoregressive rollout as a causal finite-horizon map and remove its decoded-state feedback pathway, reducing sequential model evaluation while exposing the full trajectory to planning.

    \item \textbf{Accurate and efficient long-horizon planning.}
    Across four visual-control tasks, PPWM achieves the lowest long-horizon prediction error and strongest CEM simulator performance, with over $3\times$ lower average planning latency than autoregressive LeWM.
\end{itemize}

\section{Related Work}
\label{sec:related}

\paragraph{World models for planning.}
World models support control by coupling learned dynamics with test-time optimization.
PlaNet plans in latent space from pixels~\citep{hafner2019planet}.
TD-MPC and TD-MPC2 combine learned dynamics with local trajectory optimization~\citep{hansen2022tdmpc,hansen2024tdmpc2}, while IRIS, DINO-WM, PLDM, and LeWorldModel extend planning to expressive sequence models, offline visual data, and predictive representations~\citep{micheli2023iris,zhou2025dinowm,sobal2025pldm,maes2026leworldmodel}.
However, most of these models still expose long-horizon futures through recursively composed local transitions, making both prediction and planning increasingly dependent on rollout depth.

\paragraph{Multi-step and parallel predictive dynamics.}
Long-horizon prediction can be improved either through multi-step objectives or by changing the predictive interface.
Latent overshooting improves consistency while retaining recursive dynamics~\citep{hafner2019planet}, whereas Temporal Segment Models directly parameterize future trajectory segments~\citep{mishra2017temporal}.
Direct multi-step prediction can reduce recursive bias under partial observability or misspecification, while well-specified one-step models may remain statistically preferable~\citep{somalwar2025imperfect,somalwar2026statistical}.
Most closely related, \fastlewm{} accumulates action effects with causal action prefixes and predicts multiple horizons in parallel~\citep{gao2026fastlewm}.
Existing approaches therefore either retain recursive dynamics or remove state recursion without explicitly modeling causal interaction among future representations.

\paragraph{Planning and trajectory optimization with learned world models.}
Sampling-based methods such as CEM, MPPI, and PETS optimize actions through repeated model evaluation~\citep{deboer2005cem,williams2017mppi,chua2018pets}, while differentiable planners exploit learned-model geometry or trajectory-level objectives for local refinement~\citep{georgiev2025pwm,pham2026leap,huang2026trajlewm}.
Lifted methods instead optimize intermediate states explicitly: LatCo applies latent-space collocation~\citep{rybkin2021latco}, GRASP exposes parallel optimization with virtual states~\citep{psenka2026grasp}, and GVP-WM jointly optimizes latent states and actions~\citep{ziakas2026gvpwm}.
Related control work uses direct multi-step prediction in structured state--control optimization~\citep{wu2026koopmanboxqp}.
These planners improve trajectory optimization, but their computation and optimization structure remain constrained by the temporal factorization exposed by the underlying world model.
\section{Problem Formulation}
\label{sec:problem}

We consider goal-conditioned control from visual observations.
Let $\vo_t$ denote the observation, $\va_t\in\sA$ the action, and
$\vz_t=E(\vo_t)\in\R^D$ the learned representation used for prediction and planning.
We denote by $\mathcal C_t$ the context available to the world model at time $t$,
such as recent latent states and executed actions.
Given a goal latent $\vz_g=E(\vo_g)$ and a planning horizon $H$, let
\begin{equation}
\mA=(\va_t,\ldots,\va_{t+H-1}),
\qquad
\mZ=(\vz_{t+1},\ldots,\vz_{t+H})
\end{equation}
denote a candidate action sequence and its corresponding future trajectory.
A world model predicts
\begin{equation}
\widehat{\mZ}
=
\mathcal P_\theta(\mathcal C_t,\mA),
\label{eq:predictive_model}
\end{equation}
and the planner seeks
\begin{equation}
\mA^\star
=
\arg\min_{\mA\in\sA^H}
J_{\mathrm{task}}
\left(
\mathcal P_\theta(\mathcal C_t,\mA),
\mA;
\vz_g
\right).
\label{eq:planning_objective}
\end{equation}

A conventional autoregressive world model predicts future states recursively.
Longer-horizon predictions are obtained by repeatedly feeding predicted states
back into the model,
\begin{equation}
{\widehat{\vz}_{t+j+1}=f_\theta(\widehat{\vz}_{t+j},\va_{t+j}),\qquad \widehat{\vz}_t=\vz_t,\qquad j=0,\ldots,H-1.}
\label{eq:ar_rollout}
\end{equation}
This recursive rollout has two important consequences for long-horizon planning.
First, predicted states are repeatedly reused as inputs, so local prediction errors
can propagate over the rollout.
Second, each future state depends on the preceding prediction, forming a chain of
$H$ sequentially dependent transition steps and limiting parallelism within each
trajectory.

\section{Parallel Predictive World Models for \texorpdfstring{Long-Horizon}{Long-Horizon} Planning}
\label{sec:ppwm}

PPWM is designed for long-horizon prediction and planning by exposing a causally structured finite-horizon trajectory in a single forward pass, avoiding recursive decoded-state feedback and the horizon-length sequential dependency of autoregressive rollout.
Figure~\ref{fig:overview} compares this design with autoregressive rollout and Fast-LeWM-style prefix-conditioned direct prediction.
Each prediction horizon is represented by a query conditioned on its causal action prefix, and the future queries interact through a shared causal trajectory predictor.
The same finite-horizon interface exposes all future variables jointly, which also enables lightweight trajectory-wide refinement.

\subsection{Parallel Future-State Prediction}
\label{sec:ppwm_interface}

Given the current context $\mathcal C_t$ and an action span of length $K$, PPWM predicts
\begin{equation}
    \Phi_\theta\!\left(\mathcal C_t,\mA_{t:t+K-1}\right)
    =
    \left(\widehat{\vz}_{t+1},\ldots,\widehat{\vz}_{t+K}\right).
    \label{eq:ppwm_interface}
\end{equation}
All $K$ future states are produced in a single forward pass. Unlike autoregressive rollout, $\widehat{\vz}_{t+k}$ is not fed back as the input used to generate $\widehat{\vz}_{t+k+1}$. Instead, each horizon is conditioned on the actions that causally precede it, while temporal dependence among future states is retained through hidden representations.

For each future horizon, a shared causal prefix encoder produces an action-prefix representation
\begin{equation}
    (\vp_{t+1},\ldots,\vp_{t+K})
    =
    E_\psi\!\left(\mathcal C_t,\va_{t:t+K-1};\mM_{\mathrm{causal}}\right),
    \label{eq:action_prefix}
\end{equation}
where $\vp_{t+k}$ depends only on the current context and actions up to $\va_{t+k-1}$. We then construct
\begin{equation}
    \vq_1=\vz_t,
    \qquad
    \vq_k=\vz_t+\ve_k+\mW_p\vp_{t+k},
    \qquad k=2,\ldots,K,
    \label{eq:future_queries}
\end{equation}
where $\ve_k$ identifies the prediction horizon and $\mW_p\vp_{t+k}$ represents the accumulated effect of the causal action prefix.

\subsection{Causal Trajectory Construction}
\label{sec:ppwm_causal}

The future queries are arranged in temporal order and jointly processed by a shared causal trajectory predictor.
Let $\vu_{t+k}^{(0)}=\vq_k$ denote the input representation at horizon $k$.
The predictor contains $L$ causal Transformer layers, and each layer updates the representation at horizon $k$ using only the current and preceding future positions.
\begin{equation}
\vu_{t+k}^{(\ell)}
=
\mathcal B_{\theta}^{(\ell)}
\left(
\mathcal C_t,
\vu_{t+1:t+k}^{(\ell-1)},
\mA_{t:t+k-1}
\right),
\qquad
k=1,\ldots,K,
\label{eq:causal_future_layers}
\end{equation}
where $\mathcal B_{\theta}^{(\ell)}$ denotes the $\ell$-th causal Transformer block.
The causal structure allows information from earlier future horizons to be progressively propagated and refined through the Transformer layers, while preventing access to later future positions or actions.

After the final layer, each future state is decoded from its corresponding representation.
\begin{equation}
\widehat{\vz}_{t+k}
=
F_\theta\left(\vu_{t+k}^{(L)}\right),
\qquad
k=1,\ldots,K.
\label{eq:parallel_prediction}
\end{equation}
The decoded state $\widehat{\vz}_{t+k}$ is never fed back into the prediction of later horizons, so temporal dependence is carried through the causal hidden representations rather than through recursive state prediction.

This factorization distinguishes PPWM from Fast-LeWM-style prefix-conditioned direct prediction.
As illustrated in Figure~\ref{fig:overview}, both approaches remove recursive feedback of predicted states, but PPWM additionally preserves explicit interaction among future-state representations before decoding.
PPWM therefore replaces autoregressive state feedback with causal interaction in latent space, retaining temporal structure across the predicted trajectory without restoring serial state recursion.

\subsection{Self-Conditioned Long-Horizon Composition}
\label{sec:ppwm_mar}

As our focus is the parallel predictive dynamics, we formulate the training objective directly in latent space and treat representation learning as orthogonal.
All outputs within a span are directly supervised by their corresponding future latent targets,
\begin{equation}
\mathcal L_{\mathrm{direct}}
=
\frac{1}{KD}
\sum_{k=1}^{K}
\left\|\widehat{\vz}_{t+k}-\vz^{\mathrm{GT}}_{t+k}\right\|_2^2.
\label{eq:direct_loss}
\end{equation}
For horizons longer than $K$, multiple PPWM spans are composed, so later spans receive model-generated rather than observed contexts.
Training only from observed contexts would therefore create a mismatch between single-span training and long-horizon prediction.
We address this with multi-anchor self-conditioned rollout (MAR) training.
For a sampled restart anchor $s$, let $\widehat{\mathcal C}_{t+s}$ denote the detached predicted context and let $R_s$ be the remaining supervised horizon.
We define
\begin{equation}
\mathcal L_{\mathrm{MAR}}
=
\mathbb E_s\!\left[
\frac{1}{R_sD}
\sum_{j=1}^{R_s}
\left\|
\Phi_{\theta,j}\!\left(\sg[\widehat{\mathcal C}_{t+s}],\mA_{t+s:t+s+R_s-1}\right)
-\vz^{\mathrm{GT}}_{t+s+j}
\right\|_2^2
\right],
\label{eq:mar_loss}
\end{equation}
where $\Phi_{\theta,j}$ denotes the $j$-th prediction of the restarted span and $\sg[\cdot]$ denotes the stop-gradient operator.
The final training objective is
\begin{equation}
\mathcal L_{\mathrm{PPWM}}
=
\mathcal L_{\mathrm{direct}}+\lambda_{\mathrm{MAR}}\mathcal L_{\mathrm{MAR}}.
\label{eq:ppwm_loss}
\end{equation}
The same finite-span predictor is therefore trained on the model-generated contexts encountered when spans are composed at test time.
For a horizon $H>K$, prediction remains parallel within each span and becomes sequential only across span boundaries.
\subsection{Parallel Trajectory Refinement}
\label{sec:planning}

The same finite-horizon interface can also be differentiated at trajectory level.
Because PPWM exposes the complete predicted trajectory $\widehat{\mZ}(\mA)=\Phi_\theta(\mathcal C_t,\mA)$ in one computational graph, the planner can optimize all future action and latent variables jointly instead of propagating updates through a decoded-state rollout.
We use this property as a lightweight refinement of a plan produced by the Cross-Entropy Method (CEM).

Starting from the CEM plan $\mA_0$, we introduce an auxiliary latent trajectory $\mZ$ and initialize
\begin{equation}
\mA^{(0)}=\mA_0,
\qquad
\mZ^{(0)}=\widehat{\mZ}(\mA_0).
\label{eq:trajectory_refinement_init}
\end{equation}
The refinement couples the free latent trajectory to the trajectory induced by the current actions through
\begin{equation}
\mathcal L_{\mathrm{ref}}(\mA,\mZ)
=
J_{\mathrm{task}}(\mZ)
+\alpha J_{\mathrm{task}}(\widehat{\mZ}(\mA))
+\lambda_{\mathrm{dyn}}\|\mZ-\widehat{\mZ}(\mA)\|_F^2
+\lambda_A\|\mA-\mA_0\|_F^2.
\label{eq:trajectory_refinement}
\end{equation}
The first term is the task objective evaluated on the latent trajectory $\mZ$, providing a direct optimization path toward the task goal in trajectory space.
The executable-trajectory and consistency terms keep the update tied to model-predicted dynamics, while the final term regularizes actions around the CEM solution.

At refinement step $m$, gradients with respect to the complete action and latent trajectories are computed from the same objective and updated jointly,
\begin{equation}
\mA^{(m+1)}
=
\Pi_{\mathcal B}\!\left(
\mA^{(m)}
-\eta_A\nabla_{\mA}\mathcal L_{\mathrm{ref}}\!\left(\mA^{(m)},\mZ^{(m)}\right)
\right),
\mZ^{(m+1)}
=
\mZ^{(m)}
-\eta_Z\nabla_{\mZ}\mathcal L_{\mathrm{ref}}\!\left(\mA^{(m)},\mZ^{(m)}\right).
\label{eq:trajectory_refinement_update}
\end{equation}
where $\Pi_{\mathcal B}$ enforces a local feasible action region $\mathcal B$ around the CEM plan.
Since PPWM produces all future positions within the span jointly, each refinement step updates the full action and latent trajectories without restoring a decoded-state rollout.
After a small number of refinement steps, the action candidates are re-evaluated through PPWM and selected using the executable model trajectory,
\begin{equation}
\mA^\star
=
\arg\min_{\mA\in\{\mA_0,\mA^{(1)},\ldots,\mA^{(R)}\}}
J_{\mathrm{task}}\!\left(\widehat{\mZ}(\mA)\right).
\label{eq:trajectory_refinement_selection}
\end{equation}
Keeping $\mA_0$ in the candidate set provides a fallback under the fixed model score.

\section{Analysis of Temporal Factorization}
\label{sec:analysis}

PPWM preserves temporal dependence while changing its computational factorization.
We first show that a finite autoregressive rollout is itself a causal trajectory map, so decoded-state recursion is a computational factorization rather than a requirement for temporal causality.
We then characterize the decoded-state feedback pathway that is present in autoregressive rollout but absent from PPWM, while causal interaction among future hidden representations is retained.

\subsection{Autoregressive Rollout as a Causal Trajectory Map}
\label{sec:causal_map_analysis}

Consider a deterministic autoregressive predictor over a finite horizon $K$,
\begin{equation}
\widehat{\vz}_{t+1}=g_1(\mathcal C_t,\va_t),
\qquad
\widehat{\vz}_{t+k}=g_k\!\left(\mathcal C_t,\widehat{\vz}_{t+1:t+k-1},\va_{t:t+k-1}\right),
\quad k=2,\ldots,K.
\label{eq:general_ar_factorization}
\end{equation}
Although this computation is expressed recursively, its finite-horizon input--output map need not be written in recursive form.

\begin{theorem}[Equivalent causal trajectory factorization]
\label{thm:causal_trajectory_factorization}
For every finite deterministic autoregressive predictor of the form in Eq.~(\ref{eq:general_ar_factorization}), there exist causal functions $G_1,\ldots,G_K$ such that
\begin{equation}
\widehat{\vz}_{t+k}=G_k\!\left(\mathcal C_t,\va_{t:t+k-1}\right),
\qquad
k=1,\ldots,K.
\label{eq:causal_trajectory_map}
\end{equation}
Hence the complete autoregressive rollout is exactly a causal finite-horizon map from the current context and action sequence to the predicted trajectory.
\end{theorem}

Theorem~\ref{thm:causal_trajectory_factorization} separates temporal causality from a particular computational factorization.
It shows that decoded-state recursion is not required by the causal input--output structure itself. 
PPWM realizes this structure while carrying cross-horizon dependence through causal hidden representations.

\subsection{Decoded-State Feedback Propagation}
\label{sec:decoded_feedback_analysis}

The equivalent input--output form in Theorem~\ref{thm:causal_trajectory_factorization} does not imply an equivalent computational graph.
For the canonical autoregressive transition
\begin{equation}
\widehat{\vz}_{t+m+1}=f_{\theta,m}(\widehat{\vz}_{t+m},\va_{t+m}),
\label{eq:analysis_ar_transition}
\end{equation}
every decoded prediction is reused as the input to the next transition.
PPWM instead performs causal interaction among future hidden representations and decodes each future latent only after this interaction.

\begin{theorem}[Elimination of recursive decoded-state feedback]
\label{thm:decoded_feedback}
Assume the autoregressive transitions in Eq.~(\ref{eq:analysis_ar_transition}) are differentiable, and fix the action sequence.
If the decoded state at horizon $j$ is perturbed as $\widehat{\vz}_{t+j}\leftarrow\widehat{\vz}_{t+j}+\delta$, then for any $k>j$ the induced autoregressive perturbation satisfies
\begin{equation}
\Delta\widehat{\vz}_{t+k}
=
J_{k-1}\cdots J_j\,\delta
+o(\|\delta\|),
\qquad
J_m=\frac{\partial f_{\theta,m}}{\partial \vz}\Big|_{(\widehat{\vz}_{t+m},\va_{t+m})}.
\label{eq:ar_feedback_propagation}
\end{equation}
Within a PPWM prediction span, if an additive perturbation is applied only to the decoded output node $\widehat{\vz}_{t+j}$ after all hidden future representations have been computed, then every later decoded output $\widehat{\vz}_{t+k}$, $k>j$, is unchanged.
\end{theorem}

Theorem~\ref{thm:decoded_feedback} isolates one structural error-propagation path: autoregressive rollout contains a directed decoded-state feedback chain whose local sensitivities multiply with horizon, whereas PPWM has no such path within a span.
Hidden-state perturbations may still propagate through Eq.~(\ref{eq:causal_future_layers}). This retained causal interaction is precisely what distinguishes PPWM from independent horizon-wise prediction.

\section{Experiments}
\label{sec:experiments}

We evaluate PPWM from predictive modeling to downstream planning through five questions:
\begin{itemize}
\item \textbf{Q1: Predictive fidelity.} Does parallel causal prediction improve long-horizon accuracy?
\item \textbf{Q2: Planning utility.} Does the PPWM interface improve simulator planning?
\item \textbf{Q3: Planning efficiency.} How do prediction and planning costs scale with horizon?
\item \textbf{Q4: Local refinement.} Can the parallel trajectory interface support effective refinement with feedback?
\item \textbf{Q5: Architectural generality.} Does the same parallel prediction mechanism remain effective on different world-model architectures?
\end{itemize}
Experiments use Two-Room, Reacher, Push-T, and OGBench-Cube~\citep{maes2026leworldmodel,gao2026fastlewm}.
LeWM-based models provide the main controlled comparison, while DINO-WM is used to evaluate architectural transfer.
Planning success is measured from simulator execution. Detailed protocols, hyperparameter settings, and additional ablations are provided in Appendices~\ref{app:implementation}--\ref{app:remaining}.

\subsection{Long-Horizon Prediction}
\label{sec:dynamics_comparison}

We first test whether PPWM improves long-horizon predictive accuracy.
All methods use the same frozen LeWM encoder and projector, placing predictions and targets in a shared $D=192$ latent space and making raw latent MSE directly comparable.
We compare the native autoregressive LeWM predictor, recurrent and feed-forward one-step adaptations inspired by PLDM and TD-MPC2~\citep{sobal2025pldm,hansen2024tdmpc2}, Fast-LeWM~\citep{gao2026fastlewm}, and PP-LeWM.
PP-LeWM instantiates PPWM on the LeWM architecture by retaining its representation and conditional Transformer backbone while replacing autoregressive future-state prediction with the proposed parallel causal trajectory interface.

\begin{table}[tbp]
\caption{Controlled prediction in the shared frozen LeWM latent space. Mean MSE over the first 5, 10, or 15 prediction steps (lower is better).}
\label{tab:prediction}
\centering
\resizebox{\linewidth}{!}{%
\begin{tabular}{@{}lcccccccccccc@{}}
\toprule
& \multicolumn{3}{c}{OGBench-Cube} & \multicolumn{3}{c}{Push-T} & \multicolumn{3}{c}{Two-Room} & \multicolumn{3}{c}{Reacher} \\
\cmidrule(lr){2-4}
\cmidrule(lr){5-7}
\cmidrule(lr){8-10}
\cmidrule(l){11-13}
Model
& $\bar E_{1:5}$ & $\bar E_{1:10}$ & $\bar E_{1:15}$
& $\bar E_{1:5}$ & $\bar E_{1:10}$ & $\bar E_{1:15}$
& $\bar E_{1:5}$ & $\bar E_{1:10}$ & $\bar E_{1:15}$
& $\bar E_{1:5}$ & $\bar E_{1:10}$ & $\bar E_{1:15}$ \\
\midrule
TD-MPC2
& 0.0310 & 0.0452 & 0.0525
& 0.0511 & 0.0888 & 0.1454
& 0.2126 & 0.2601 & 0.2951
& 0.0628 & 0.0670 & 0.0703 \\
PLDM
& 0.0200 & 0.0325 & 0.0374
& 0.0371 & 0.0740 & 0.1319
& 0.0749 & 0.1153 & 0.1480
& 0.0557 & 0.0586 & 0.0605 \\
LeWM
& 0.0173 & 0.0343 & 0.0456
& 0.0312 & 0.0967 & 0.1852
& 0.1139 & 0.1867 & 0.2707
& 0.0133 & 0.0199 & 0.0266 \\
Fast-LeWM
& 0.0144 & 0.0268 & 0.0319
& 0.0189 & 0.0526 & 0.1047
& 0.0536 & 0.1106 & 0.1610
& 0.0520 & 0.0555 & 0.0728 \\
\midrule
\textbf{PP-LeWM}
& \textbf{0.0130} & \textbf{0.0222} & \textbf{0.0270}
& \textbf{0.0187} & \textbf{0.0485} & \textbf{0.0949}
& \textbf{0.0395} & \textbf{0.0780} & \textbf{0.1159}
& \textbf{0.0007} & \textbf{0.0009} & \textbf{0.0011} \\
\bottomrule
\end{tabular}%
}
\end{table}

\begin{figure}[tbp]
\centering
\includegraphics[width=0.98\textwidth]{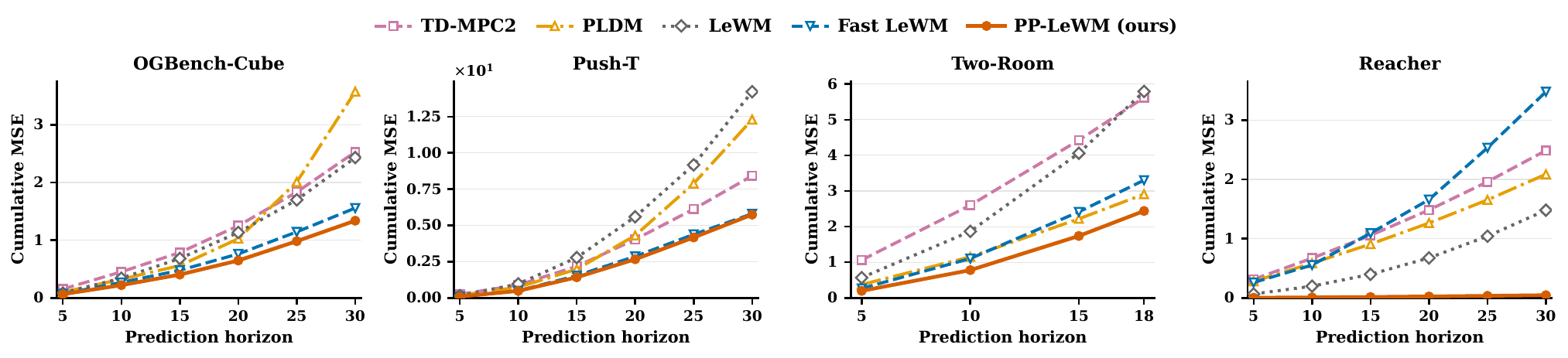}
\caption{Cumulative latent prediction MSE as the open-loop prediction horizon increases. All methods use the same frozen LeWM representation (lower is better).}
\label{fig:prediction_error_curves}
\end{figure}

Table~\ref{tab:prediction} shows that PP-LeWM achieves the lowest MSE at all reported horizons on all four tasks.
Fast-LeWM is the closest direct parallel baseline. 
It removes decoded-state recursion through action-prefix-conditioned prediction, whereas PP-LeWM additionally retains causal interaction among future representations before decoding.
Because Fast-LeWM already removes decoded-state recursion, the consistent gap points to cross-horizon interaction as a useful design choice inside a parallel predictor.
Reacher shows the largest separation. Its smooth, low-dimensional dynamics provide a regime in which cross-horizon structure can be exploited consistently across the prediction span.
The cumulative-error curves in Figure~\ref{fig:prediction_error_curves} retain the same ordering beyond the tabulated horizons.
PP-LeWM exhibits the slowest cumulative error growth across all four tasks, while the gap to autoregressive LeWM and Fast-LeWM generally widens at longer horizons.
The separation grows most clearly on Reacher and Two-Room, where PP-LeWM accumulates substantially less error as prediction moves farther from the observed context.


\subsection{Planning Across Predictive Interfaces}
\label{sec:controlled_cem}

We evaluate whether the predictive differences persist after coupling each model to downstream planners.
We compare LeWM, Fast-LeWM, and PP-LeWM under trajectory-wide LatCo and GRASP optimization as well as population-based CEM planning.
Each completed trial plans over a horizon of 50 primitive simulator actions, and executes the full plan open loop.
Anytime and Terminal success are computed only from the resulting simulator trajectory. Table~\ref{tab:same_cem} summarizes the resulting open-loop success rates across planners and predictive interfaces.

Under CEM, PP-LeWM achieves the highest anytime and terminal success on all four tasks.
The lifted planners expose a different model--planner interaction: Fast-LeWM remains competitive on Cube and Two-Room but degrades sharply with LatCo on Push-T and Reacher, while GRASP exhibits high anytime but low terminal success on Reacher.
CEM evaluates Fast-LeWM through its native direct finite-horizon interface, whereas the LatCo and GRASP adaptations require a chained one-step dynamics factor over optimized latent states.
Appendix~\ref{app:lifted_realizability_diag} shows large virtual-to-rollout gaps when Fast-LeWM is forced into this chained local factor.
The mismatch is therefore an interface issue: lifted optimization can improve the local surrogate without improving the native finite-horizon prediction used by the direct model.

\begin{table}[tbp]
\caption{Open-loop simulator success across predictive interfaces (100 episodes $\times$ 3 planner seeds; higher is better). CEM uses each model's frozen scoring rule; see Appendix~\ref{app:pp_tail_cem_protocol}.}
\label{tab:same_cem}
\centering
\resizebox{\linewidth}{!}{%
\begin{tabular}{@{}llcc cc cc cc@{}}
\toprule
& & \multicolumn{2}{c}{OGBench-Cube} & \multicolumn{2}{c}{Push-T} & \multicolumn{2}{c}{Two-Room} & \multicolumn{2}{c}{Reacher} \\
\cmidrule(lr){3-4}
\cmidrule(lr){5-6}
\cmidrule(lr){7-8}
\cmidrule(l){9-10}
Planner & Model
& Anytime$\uparrow$ & Terminal$\uparrow$
& Anytime$\uparrow$ & Terminal$\uparrow$
& Anytime$\uparrow$ & Terminal$\uparrow$
& Anytime$\uparrow$ & Terminal$\uparrow$ \\
\midrule
\multirow{3}{*}{LatCo}
& LeWM
& 28.67 & 20.67
& 0.67 & 0.33
& 3.33 & 1.67
& 35.67 & 18.67 \\
& Fast-LeWM
& 31.00 & 22.33
& 1.67 & 0.67
& 7.67 & 5.67
& 15.67 & 2.33 \\
& PP-LeWM
& 31.00 & 26.00
& 16.67 & 12.67
& 8.67 & 5.33
& 79.33 & 78.67 \\
\midrule
\multirow{3}{*}{GRASP}
& LeWM
& 31.33 & 21.67
& 9.00 & 0.33
& 9.33 & 5.00
& 81.00 & 11.33 \\
& Fast-LeWM
& 29.00 & 18.67
& 15.00 & 2.00
& 24.00 & 20.00
& 62.67 & 5.67 \\
& PP-LeWM
& 34.00 & 26.00
& 17.00 & 3.67
& 10.33 & 4.67
& 38.67 & 0.67 \\
\midrule
\multirow{3}{*}{CEM}
& LeWM
& 35.00 & 28.67
& 35.33 & 27.67
& 49.67 & 49.67
& 76.67 & 64.33 \\
& Fast-LeWM
& 38.00 & 35.33
& 35.33 & 28.67
& 44.67 & 44.33
& 51.33 & 35.00 \\
& \textbf{PP-LeWM}
& \textbf{45.67} & \textbf{39.33}
& \textbf{37.67} & \textbf{29.33}
& \textbf{55.33} & \textbf{55.00}
& \textbf{97.00} & \textbf{96.67} \\
\bottomrule
\end{tabular}%
}
\end{table}

\subsection{Planning Efficiency and Horizon Scaling}
\label{sec:planning_efficiency}

\begin{figure}[tbp]
\centering
\begin{minipage}[t]{0.475\textwidth}
\vspace{0pt}
\centering
\captionof{table}{CEM planning latency at $H{=}10$ (ms); speedup relative to LeWM is shown below.}
\resizebox{0.95\linewidth}{!}{%
\begin{tabular}{@{}lccc@{}}
\toprule
Dataset & LeWM & Fast-LeWM & PP-LeWM \\
\midrule
OGBench-Cube & \shortstack{1245.4\\$1.00\times$} & \shortstack{474.3\\$2.63\times$} & \shortstack{\textbf{386.7}\\\textbf{$3.22\times$}} \\
Push-T & \shortstack{1252.9\\$1.00\times$} & \shortstack{457.7\\$2.74\times$} & \shortstack{\textbf{357.0}\\\textbf{$3.51\times$}} \\
Two-Room & \shortstack{1243.8\\$1.00\times$} & \shortstack{505.9\\$2.46\times$} & \shortstack{\textbf{385.6}\\\textbf{$3.23\times$}} \\
Reacher & \shortstack{1198.6\\$1.00\times$} & \shortstack{465.3\\$2.58\times$} & \shortstack{\textbf{368.9}\\\textbf{$3.25\times$}} \\
\midrule
Average & \shortstack{1235.2\\$1.00\times$} & \shortstack{475.8\\$2.60\times$} & \shortstack{\textbf{374.5}\\\textbf{$3.30\times$}} \\
\bottomrule
\end{tabular}%
}
\label{tab:planning_latency_10step}
\end{minipage}
\hfill
\begin{minipage}[t]{0.50\textwidth}
\vspace{0pt}
\centering
\includegraphics[width=\linewidth]{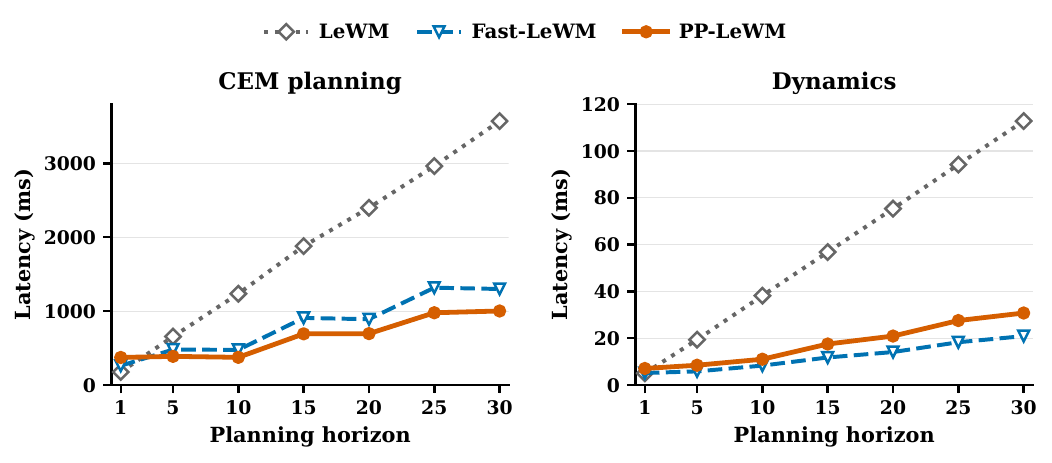}
\captionof{figure}{Average CEM planning and dynamics latency across four tasks. }
\label{fig:latency_scaling}
\end{minipage}
\end{figure}

Planning efficiency is evaluated by measuring how dynamics and end-to-end CEM latency scale.
Table~\ref{tab:planning_latency_10step} summarizes CEM planning latency at a ten-step model horizon, while Figure~\ref{fig:latency_scaling} reports average CEM planning and dynamics latency for planning horizons of 1, 5, 10, 15, 20, 25, and 30 steps.
PP-LeWM achieves the lowest CEM planning latency on all four tasks, averaging 374.5\,ms compared with 1235.2\,ms for LeWM and 475.8\,ms for Fast-LeWM.
This corresponds to a $3.30\times$ average speedup over LeWM.
Figure~\ref{fig:latency_scaling} further shows that LeWM latency grows rapidly with rollout length, whereas PP-LeWM exhibits span-wise growth and remains substantially more efficient at medium and long horizons.
Fast-LeWM has the lowest raw dynamics latency due to its lighter direct predictor, but its re-anchored self-consistency score adds planner-side model evaluations. Score-level details are provided in Appendix~\ref{app:cem_score_control}.
PP-LeWM therefore delivers the lowest end-to-end CEM latency while retaining causal interaction among future representations.

\subsection{Closed-Loop Trajectory Refinement}
\label{sec:trajectory_optimization}

To evaluate trajectory-level refinement under feedback, we run PP-LeWM in closed loop on Push-T and Two-Room.
The planner replans every five primitive simulator steps.
At each decision, the agent receives a fresh observation, runs CEM over the remaining episode horizon, applies the local refinement, executes the first five primitive actions, and replans.
We compare the unrefined CEM plan with action-only Adam and trajectory-wide refinement, which jointly updates the action sequence and an auxiliary latent trajectory.
All Anytime and Terminal success rates are computed exclusively from the resulting closed-loop simulator trajectories, as shown in Table~\ref{tab:planner_comparison}.

Trajectory refinement provides the clearest closed-loop gain on Two-Room, improving both Anytime and Terminal success by 2.67 percentage points over CEM and consistently outperforming action-only Adam.
The additional gain over action-only Adam on Two-Room comes with the ability to move the latent trajectory and actions together across the full horizon.
Moreover, the parallel trajectory interface reduces the sequential prediction depth of each refinement iteration from $\mathcal{O}(H)$ autoregressive transitions to $\mathcal{O}(\lceil H/K\rceil)$ PPWM span evaluations, and to $\mathcal{O}(1)$ when $H\leq K$.
Closed-loop replanning therefore combines trajectory-wide updates with a short sequential optimization path while repeatedly re-anchoring the model to observed states.

\subsection{Architectural Generality on DINO-WM}
\label{sec:dino_transfer}

To test architectural transfer, we instantiate PPWM on DINO-WM~\citep{zhou2025dinowm}.
PP-DINO-WM retains the frozen DINOv2 representation~\citep{oquab2023dinov2} while replacing autoregressive future prediction with the proposed parallel causal trajectory interface.
The autoregressive DINO-WM and PP-DINO-WM use the same test windows, histories, and future actions and are rolled out open loop. Table~\ref{tab:dino_transfer} reports prediction error across the evaluated horizons.
PP-DINO-WM reduces visual latent error at every reported horizon, with the relative reduction increasing from 18.84\% over 5 steps to 36.45\% over 20 steps.
The same temporal factorization therefore remains effective after replacing the LeWM backbone with DINO-WM.

\begin{table}[t]
\centering
\begin{minipage}[t]{0.49\textwidth}
\vspace{0pt}
\centering
\captionof{table}{Closed-loop refinement with PP-LeWM under a shared CEM planner (higher is better).}
\label{tab:planner_comparison}
\resizebox{\linewidth}{!}{%
\begin{tabular}{@{}lcc cc@{}}
\toprule
& \multicolumn{2}{c}{Push-T} & \multicolumn{2}{c}{Two-Room} \\
\cmidrule(lr){2-3}
\cmidrule(l){4-5}
Method
& Anytime$\uparrow$ & Terminal$\uparrow$
& Anytime$\uparrow$ & Terminal$\uparrow$ \\
\midrule
CEM
& 56.33 & 49.00
& 55.33 & 55.00 \\
+ Adam
& 57.00 & \textbf{50.33}
& 56.00 & 56.00 \\
\textbf{+ Traj. Refine}
& \textbf{58.33} & 50.00
& \textbf{58.00} & \textbf{57.67} \\
\bottomrule
\end{tabular}%
}
\end{minipage}
\hfill
\begin{minipage}[t]{0.49\textwidth}
\vspace{0pt}
\centering
\captionof{table}{PP-DINO-WM transfer on Push-T. DINOv2 visual-patch MSE (lower is better).}
\label{tab:dino_transfer}
\resizebox{\linewidth}{!}{%
\begin{tabular}{@{}lccc@{}}
\toprule
Prediction Steps & DINO-WM & PP-DINO-WM & Reduction \\
\midrule
5  & 0.098855 & \textbf{0.080229} & 18.84\% \\
10 & 0.154497 & \textbf{0.108464} & 29.80\% \\
15 & 0.219716 & \textbf{0.147109} & 33.05\% \\
20 & 0.284458 & \textbf{0.180760} & 36.45\% \\
\bottomrule
\end{tabular}%
}
\end{minipage}
\end{table}

\subsection{PPWM Component Analysis}
\label{sec:component_analysis}

Table~\ref{tab:mar_ablation} isolates the two main components of PP-LeWM on Push-T.
Fast-LeWM provides a parallel reference without causal interaction among future representations, while PP-LeWM w/o MAR removes multi-anchor rollout training.
Full PP-LeWM achieves the lowest error at every horizon.
These results show that causal trajectory construction and MAR are both important for maintaining accurate long-horizon prediction.

\begin{table}[tbp]
\caption{Push-T component analysis using mean latent MSE; lower is better.}
\label{tab:mar_ablation}
\centering
\resizebox{0.6\linewidth}{!}{%
\begin{tabular}{@{}lcccc@{}}
\toprule
Model & $\bar E_{1:10}\downarrow$ & $\bar E_{1:15}\downarrow$ & $\bar E_{1:20}\downarrow$ & $\bar E_{1:30}\downarrow$ \\
\midrule
Fast-LeWM & 0.052634 & 0.104680 & 0.146183 & 0.208614 \\
PP-LeWM w/o MAR & 0.051014 & 0.102552 & 0.150012 & 0.239441 \\
\textbf{PP-LeWM} & \textbf{0.048525} & \textbf{0.094867} & \textbf{0.133276} & \textbf{0.191456} \\
\bottomrule
\end{tabular}%
}
\end{table}

\section{Conclusion}

We introduced Parallel Predictive World Models (PPWM) for accurate and efficient long-horizon planning.
PPWM replaces recursive decoded-state prediction with parallel causal trajectory prediction, preserving temporal interaction among future representations while exposing the finite-horizon trajectory jointly to planning.
Across four visual-control tasks, PPWM consistently improves long-horizon prediction and CEM planning performance while achieving over a $3\times$ average planning speedup relative to autoregressive LeWM.
The same factorization transfers to DINO-WM and naturally supports trajectory-wide refinement, demonstrating its compatibility with different world-model backbones and planning regimes.
More broadly, PPWM shows that long-horizon world models can preserve causal temporal structure without relying on state-by-state autoregressive rollout, providing a direct and efficient alternative for trajectory prediction and planning.

\subsection*{AI use statement}
Generative AI tools were used for language editing, manuscript organization, literature search, reference verification, and notation and consistency checking. All AI-assisted suggestions and bibliographic information were manually verified by the authors. The authors take full responsibility for the final manuscript.

\subsection*{Reproducibility statement}
Sections~\ref{sec:problem}--\ref{sec:experiments} define PPWM, its predictive and planning formulations, and the evaluation questions. Appendix~\ref{app:implementation} provides implementation details, Appendix~\ref{app:theory} states assumptions and derivations, Appendix~\ref{app:protocol} specifies the experimental protocols and metric definitions, and Appendix~\ref{app:remaining} collects additional ablations and diagnostics. The code will be made publicly available.

\bibliographystyle{plainnat}
\bibliography{references}

\appendix
\section{Implementation Details}
\label{app:implementation}

\subsection{PP-LeWM instantiation}

\paragraph{Representation, temporal scale, and context.}
The visual encoder and projector are frozen throughout PP-LeWM training. 
The latent dimension is $D=192$. 
For Push-T and Reacher, one model-step action is the same five-primitive-action block used by LeWM. With two-dimensional primitive controls, the raw block dimension is 10. The main configuration uses $K=10$ future model steps.

\paragraph{History context encoder.}
The action-prefix branch is conditioned on a compact motion context rather than only the latest latent. We compute
\begin{equation}
 u_t=[z_t,\;z_t-z_{t-1},\;z_{t-1}-z_{t-2}]\in\mathbb R^{576}
\end{equation}
and map it with a lightweight MLP
\begin{equation}
576\rightarrow384\rightarrow192
\end{equation}
using normalization and GELU between linear layers. Its output $h_t\in\mathbb R^{192}$ is used only by the action-prefix encoder. The original three-state LeWM trajectory input is retained separately.


\paragraph{History-conditioned action-prefix encoder.}
Each future action block is first projected to 32 dimensions and then mapped through a $32\rightarrow768\rightarrow192$ MLP.
The history context $h_t$ is separately mapped through a $192\rightarrow768\rightarrow192$ state-token MLP.
We prepend this state token to the sequence of future action tokens, add fixed sinusoidal positional encodings, and process the sequence with a three-layer causal Transformer with model dimension 192, six heads, and FFN dimension 768.
Dropping the context-token output yields $p_{t+1:t+K}\in\mathbb R^{K\times192}$.
A LayerNorm followed by a linear $192\rightarrow192$ prefix projection $W_p$ maps these features into the future-query space. The projection is initialized near zero so that the new prefix branch initially perturbs the pretrained trajectory dynamics only weakly.
\paragraph{Horizon-aware future queries.}
Fixed horizon codes $e_k\in\mathbb R^{192}$ are added to the future-query tokens. We preserve the one-step LeWM prior by setting
\begin{equation}
q_1=z_t,
\qquad
q_k=z_t+e_k+W_p p_{t+k},\quad k=2,\ldots,K.
\end{equation}
The full trajectory sequence is therefore
\begin{equation}
[z_{t-2},z_{t-1},q_1,\ldots,q_K].
\end{equation}
The causal mask permits hidden trajectory interaction from earlier to later query tokens, but no predicted latent $\hat z_{t+k}$ is fed back as the input used to generate $\hat z_{t+k+1}$ within the span.

\paragraph{Conditional dynamics backbone and prediction head.}
The base LeWM predictor contains six causal conditional Transformer blocks with hidden dimension 192, AdaLN-style local action conditioning, and a 2048-dimensional feed-forward sublayer.
The main model fine-tunes all six predictor blocks, the final normalization, and the pretrained prediction projection.
The encoder, projector, and LeWM action encoder remain frozen.
Future hidden states are decoded directly by the fine-tuned pretrained prediction projection.
\paragraph{Training curriculum.}
New history/prefix modules are first warmed up with the dynamics backbone frozen and $\cL_{\mathrm{direct}}$ only.
We then jointly fine-tune the full dynamics backbone using a smaller learning rate for pretrained dynamics parameters than for the new history/prefix modules.
After direct prediction stabilizes, MAR is enabled.
During development, grouped learning rates are selected from $10^{-4}$--$3\times10^{-4}$ for new modules and $10^{-5}$--$3\times10^{-5}$ for pretrained dynamics parameters.
\paragraph{Direct and multi-anchor rollout losses.}
All $K$ direct outputs receive equal-weight latent MSE supervision. For MAR, after the first span we choose two random anchors $s_1,s_2\sim U(3,K-1)$ and the boundary anchor $K$. At every anchor, the three predicted states are detached before the next model call. The history encoder and prefix encoder are then recomputed from this detached predicted history, so the rollout objective trains the new action representation under self-generated dynamics context without backpropagating through the first-span predicted history. The final objective is
\begin{equation}
\cL_{\mathrm{PPWM}}=\cL_{\mathrm{direct}}+0.25\,\cL_{\mathrm{MAR}}.
\end{equation}
The base configuration uses no auxiliary task-specific losses. Nominal and action-perturbed trajectories share the same objective.

\subsection{Controlled dynamics baselines}
\label{app:dynamics_baselines}

The controlled PLDM and TD-MPC2 baselines use the same cached LeWM latents and frozen action embedding as the other dynamics models, so the comparison isolates the temporal predictor.
We retain their characteristic recurrent or feed-forward transition structures while omitting agent-specific components outside the controlled dynamics comparison.

\paragraph{PLDM.}
PLDM uses a recurrent predictor in its Two-Room instantiation. The original paper reports a two-layer GRU with 512 hidden dimensions~\citep{sobal2025pldm}.
Our controlled PLDM-GRU baseline projects the current 192-dimensional LeWM latent into the GRU hidden state, consumes the frozen LeWM action embeddings recurrently, and maps each recurrent output back to the 192-dimensional LeWM latent space.
It is rolled out autoregressively at evaluation time.
This baseline tests whether recurrent memory alters the long-horizon behavior of one-step autoregressive prediction while keeping the visual representation fixed.
\paragraph{TD-MPC2.}
TD-MPC2 uses a feed-forward one-step latent transition of the form $z_{t+1}=d_\theta(z_t,a_t)$ implemented as an MLP~\citep{hansen2024tdmpc2}. Our controlled adaptation concatenates the frozen LeWM latent and action embedding, predicts the next LeWM latent with an MLP transition, and recursively feeds predictions back at evaluation time.

\paragraph{Training control.}
The two adapted baselines are trained only on the cached LeWM latent/action trajectories used for the PPWM comparison. Their objectives supervise predicted LeWM latents and do not include the representation-learning losses of their source methods. Parameter counts, optimizer schedules, and exact hidden widths follow the controlled experiment configuration.

\subsection{PP-DINO-WM instantiation}
\label{app:pp_dino_implementation}

\paragraph{Representation and temporal scale.}
We instantiate PPWM on the Push-T DINO-WM backbone while keeping the DINOv2 ViT-S/14 visual encoder frozen. RGB observations are normalized to $[-1,1]$ and resized to $196\times196$ before DINOv2 encoding. Each observation is represented by 196 spatial patch tokens of dimension 384. The four-dimensional proprioceptive state is linearly encoded to 10 dimensions and concatenated to every patch, giving a 394-dimensional state token. Histories contain three frames separated by five primitive environment steps. Five primitive two-dimensional actions are grouped into one 10-dimensional model-step action, and the native parallel span is $K=10$, corresponding to 50 primitive environment steps.

\paragraph{Parallel dynamics.}
PP-DINO-WM uses the same PPWM factorization as Section~\ref{sec:ppwm}: a causal action-prefix branch constructs horizon-specific future queries, which are jointly processed by the DINO-WM causal dynamics backbone. The prefix encoder has three Transformer layers with width 192, six attention heads, and FFN dimension 768. The DINO-WM dynamics backbone has six layers, 16 attention heads, and FFN dimension 2048. Attention is causal at the frame level, so each future frame can interact with preceding future-frame representations while remaining independent of later frames. All $K$ future states are produced in one backbone call without feeding decoded states back within the span.

\paragraph{Initialization and training.}
The dynamics predictor, action encoder, and proprioceptive encoder are initialized from the official Push-T DINO-WM checkpoint, while the history/prefix branch is newly initialized.
The episode-separated split contains 301,279 training windows, 512 development windows, and 1,008 test windows.
The first five epochs train only the new branch. From epoch 6 onward, the dynamics predictor and action encoder are jointly fine-tuned.
The results in Section~\ref{sec:dino_transfer} use the fixed epoch-10 checkpoint and therefore reflect direct 10-step supervision without MAR.
Training uses four-GPU data parallelism and AdamW with learning rates of $10^{-4}$ for newly initialized parameters and $10^{-5}$ for pretrained dynamics parameters, weight decay $10^{-4}$, gradient-norm clipping at 1, FP32 arithmetic, and an effective batch size of 128.

\subsection{Parallel Trajectory Refinement Details}
\label{app:coto_details}

\paragraph{CEM initialization and trajectory variables.}
At each closed-loop decision, refinement starts from the CEM action sequence $\mA_0$ and its PPWM rollout $\widehat{\mZ}(\mA_0)$. The variable $\mZ$ is auxiliary, while executable trajectories are always generated by the frozen PPWM.

\paragraph{Tail task cost.}
CEM and the trajectory refinement use the same normalized tail objective.
For the remaining horizon $H$, let $q=\min(4,H)$ and define
\begin{equation}
J_{\mathrm{task}}=J_\rho(\mathbf Z,\vz_g)
=
\frac{\sum_{j=0}^{q-1}\rho^j\operatorname{MSE}(\mathbf z_{H-j},\vz_g)}{\sum_{j=0}^{q-1}\rho^j}.
\label{eq:coto_tail}
\end{equation}
The terminal state receives the largest weight, and at most the final four model states contribute to the score.
The refinement objective in Eq.~(\ref{eq:trajectory_refinement}) uses this score for both the free trajectory and the executable PPWM rollout, together with a soft dynamics-consistency penalty and a quadratic action penalty around the CEM reference plan.

\section{Proofs and Additional Analysis}
\label{app:theory}

\subsection{Proof of Theorem~\ref{thm:causal_trajectory_factorization}}
\label{app:proof_causal_trajectory_factorization}

We prove the claim by recursive substitution.
For the first horizon, define
\begin{equation}
G_1(\mathcal C_t,\va_t)=g_1(\mathcal C_t,\va_t).
\end{equation}
Then $\widehat{\vz}_{t+1}=G_1(\mathcal C_t,\va_t)$.
Assume that for every $j<k$ there exists a function $G_j$ satisfying
\begin{equation}
\widehat{\vz}_{t+j}=G_j(\mathcal C_t,\va_{t:t+j-1}).
\end{equation}
Substituting these expressions into the $k$-th autoregressive prediction gives
\begin{equation}
G_k(\mathcal C_t,\va_{t:t+k-1})
=
g_k\!\left(\mathcal C_t,G_1(\mathcal C_t,\va_t),\ldots,G_{k-1}(\mathcal C_t,\va_{t:t+k-2}),\va_{t:t+k-1}\right).
\label{eq:causal_map_recursive_definition}
\end{equation}
Therefore $\widehat{\vz}_{t+k}=G_k(\mathcal C_t,\va_{t:t+k-1})$.
By induction, this construction holds for every $k=1,\ldots,K$.
Each $G_k$ depends only on the current context and the action prefix ending at $t+k-1$, so the resulting finite-horizon map is causal.

\subsection{Proof of Theorem~\ref{thm:decoded_feedback}}
\label{app:proof_decoded_feedback}

Perturb $\widehat{\vz}_{t+j}$ to $\widehat{\vz}_{t+j}+\delta$ while keeping the action sequence fixed, and let $\Delta_m$ denote the induced perturbation of $\widehat{\vz}_{t+m}$.
Differentiability of Eq.~(\ref{eq:analysis_ar_transition}) gives
\begin{equation}
\Delta_{m+1}=J_m\Delta_m+o(\|\Delta_m\|),
\label{eq:feedback_one_step}
\end{equation}
where $J_m$ is defined in Eq.~(\ref{eq:ar_feedback_propagation}).
Starting from $\Delta_j=\delta$ and recursively applying Eq.~(\ref{eq:feedback_one_step}) yields
\begin{equation}
\Delta_k
=
J_{k-1}\cdots J_j\,\delta
+o(\|\delta\|),
\end{equation}
which proves the autoregressive statement.
Hence the autoregressive computational graph contains a directed path from each decoded future state to every later decoded state.

For PPWM, the future hidden representations are computed by the causal trajectory blocks before decoding, as in Eq.~(\ref{eq:causal_future_layers}), and each output is obtained by
\begin{equation}
\widehat{\vz}_{t+k}=F_\theta(\vu_{t+k}^{(L)}).
\end{equation}
No decoded output $\widehat{\vz}_{t+j}$ is an input to any later hidden representation $\vu_{t+k}^{(\ell)}$.
Therefore, after the hidden representations have been fixed, intervening on $\widehat{\vz}_{t+j}$ after $F_\theta$ creates no directed computational path to any later output and induces exactly zero change in $\widehat{\vz}_{t+k}$ for $k>j$.
At the same time, causal hidden interaction remains because $\vu_{t+k}^{(\ell)}$ can depend on earlier future representations $\vu_{t+1:t+k}^{(\ell-1)}$.
The result therefore removes recursive decoded-state feedback without asserting independence among PPWM predictions.

\section{Experimental Protocols}
\label{app:protocol}

This appendix specifies the common evaluation protocol and the experiment-specific deviations used in the main text.

\subsection{Evaluation Setup}
\label{app:evaluation_setup}

One model step corresponds to five primitive simulator actions. Unless stated otherwise, open-loop planning uses a one-shot horizon of 10 model steps (50 primitive actions), and each configuration is evaluated on 100 episodes with three planner seeds. CEM uses 300 candidates, 30 elites, and 30 iterations in standardized action coordinates.
Anytime success records whether the native task success condition is reached at any point during simulator execution, while Terminal success evaluates the final primitive step. Planner hyperparameters are selected on development data and frozen before test evaluation.

\subsection{Controlled Prediction Protocol}
\label{app:controlled_dynamics}

The shared representation, baseline constructions, and training controls are described in Appendix~\ref{app:dynamics_baselines}. All methods receive identical initial histories, future actions, and latent targets.
For horizon $j$, we measure
\begin{equation}
E_j=\frac{1}{D}\left\|\hat z_{t+j}-z^{\mathrm{GT}}_{t+j}\right\|_2^2.
\label{eq:prediction_metric}
\end{equation}
The mean error over the first $h$ steps and the cumulative error are
\begin{equation}
\bar E_{1:h}=\frac{1}{h}\sum_{j=1}^{h}E_j,
\qquad
C_h=\sum_{j=1}^{h}E_j=h\bar E_{1:h}.
\label{eq:prediction_aggregate_metrics}
\end{equation}
Table~\ref{tab:prediction} reports $\bar E_{1:h}$ and Figure~\ref{fig:prediction_error_curves} reports $C_h$.

\subsection{Open-Loop CEM Protocol}
\label{app:pp_tail_cem_protocol}

The CEM rows in Table~\ref{tab:same_cem} follow the common setup above: each trial plans once from the observed three-frame history and goal, then executes the complete 50-step action sequence without replanning or early stopping.
Push-T success requires the joint agent/block positional criterion and block-angle error below $\pi/9$. Reacher requires both absolute joint-position errors below 0.05 radians. Two-Room uses final agent-to-goal distance below 16 pixels, and OGBench-Cube uses final cube-to-goal distance at most 0.04 meters.

\subsection{Open-Loop Trajectory Optimization Protocol}
\label{app:planner_protocol}

All planners use the same standardized action coordinates and environment bounds. LatCo uses its stochastic action initialization with start--goal state interpolation, while GRASP uses its lifted-state initialization with latent perturbations and periodic rollout synchronization. These initializations are kept separate from the CEM-initialized local refinement in Section~\ref{sec:trajectory_optimization}.

\subsection{Closed-Loop Refinement Protocol}
\label{app:closed_loop_protocol}

The closed-loop study uses the common CEM search budget, but executes only one five-action block before receiving a new simulator observation and replanning. The remaining model horizon decreases from 10 to one over the 50-step episode, and both tasks use $\rho=0.25$.
All methods start each decision from the same CEM solution and preserve the unrefined plan as a fallback candidate.
Adam refines only the standardized action sequence, whereas trajectory refinement jointly updates the action sequence and a free virtual latent trajectory.
The frozen trajectory-refinement settings are
\begin{center}
\begin{tabular}{@{}lcc@{}}
\toprule
Parameter & Push-T & Two-Room \\
\midrule
Rounds $R$ & 2 & 3 \\
$\eta_A$ & 1 & 2 \\
$\eta_Z$ & 24 & 48 \\
$\alpha$ & 1 & 1 \\
$\lambda_{\mathrm{dyn}}$ & 1 & 1 \\
$\lambda_A$ & 0.01 & 0.01 \\
Trust-region radius $\delta$ & 1 & 0.25 \\
Tail coefficient $\rho$ & 0.25 & 0.25 \\
\bottomrule
\end{tabular}
\end{center}

\subsection{Single-GPU Timing Protocol}
\label{app:scaling_diagnostics}

Timing uses horizons $H\in\{1,5,10,15,20,25,30\}$ on a single NVIDIA A100-SXM4-80GB GPU in FP32 evaluation mode.
All methods receive the same three-frame latent history, past actions, and future action sequence. Data loading, image encoding, checkpoint loading, and simulator execution are excluded. Dynamics latency measures one evaluation of 300 candidate trajectories, and end-to-end CEM timing uses the common search budget from Appendix~\ref{app:evaluation_setup}.
LeWM is rolled out autoregressively, while Fast-LeWM and PP-LeWM are composed in spans of at most 10 model steps.
A complete CEM solve includes candidate sampling, action processing, dynamics prediction, objective evaluation, sorting, elite updates, and final plan scoring.
CUDA synchronization is applied around timed regions, and the horizon-scaling curves report median wall-clock time.

\subsection{DINO-WM Architectural Generality Protocol}
\label{app:backbone_transfer}

The DINO-WM experiment evaluates both models on the same 1,008 held-out Push-T windows and known future action sequences. DINO-WM rolls out autoregressively, whereas PP-DINO-WM predicts 10 steps per span and composes two spans for the 20-step horizon.
Prediction error is computed on the 196-by-384 DINOv2 visual patch features, averaging the stepwise MSE over all valid windows at each reported horizon.

\section{Additional Ablations and Diagnostics}
\label{app:remaining}

This appendix contains three focused analyses that directly support the main results: a planner-interface diagnostic for the lifted Fast-LeWM comparisons, ablations of the PPWM parallel prediction span and self-conditioned long-horizon training, and a controlled analysis of the CEM scoring function.

\subsection{Planner-Interface Diagnostic}
\label{app:lifted_realizability_diag}

\begin{table}[H]
\caption{Planner-interface diagnostic for the Fast-LeWM lifted-planner adaptations. ``Virtual'' is the optimized terminal latent state, ``chained rollout'' re-evaluates the one-step planning factor from the true initial context, and ``simulator'' is the realized terminal latent MSE after executing the optimized actions. All entries use the completed 300-trial evaluations. Lower is better.}
\label{tab:lifted_realizability}
\centering
\setlength{\tabcolsep}{2.0pt}
\begin{tabular}{@{}llccc@{}}
\toprule
Task & Planner & Virtual MSE & Chained MSE & Simulator MSE \\
\midrule
Push-T & LatCo & 0.0036 & 1.5570 & 1.5449 \\
Push-T & GRASP & 1.0202 & 1.0155 & 1.2472 \\
Reacher & LatCo & 0.0002 & 0.9086 & 1.0736 \\
Reacher & GRASP & 0.1736 & 0.1752 & 0.6278 \\
\bottomrule
\end{tabular}
\end{table}

The large virtual-to-rollout gaps show that the chained one-step factor used by lifted Fast-LeWM can become misaligned with its native direct predictor, especially for LatCo.
Fast-LeWM consequently performs well when CEM queries its native finite-horizon predictor, but degrades when lifted optimization replaces that interface with a chained local factor.

\subsection{PPWM Component Ablations}
\label{app:ablation_protocol}

We examine three design choices tied to the proposed parallel predictor: the native parallel prediction span $K$, causal trajectory construction, and multi-anchor self-conditioned rollout training (MAR).

\paragraph{Parallel prediction span.}
For the span study, $K\in\{5,10,15\}$ while all other predictor and rollout settings are held fixed.
\begin{table}[H]
\caption{Effect of the native parallel prediction span $K$. Entries are mean latent MSE over the first $h$ prediction steps, $\bar E_{1:h}$ (lower is better).}
\label{tab:prediction_span_ablation}
\centering
\setlength{\tabcolsep}{2.0pt}
\begin{tabular}{@{}lccccc@{}}
\toprule
Dataset & $K$ & $\bar E_{1:5}\downarrow$ & $\bar E_{1:10}\downarrow$ & $\bar E_{1:15}\downarrow$ & $\bar E_{1:20}\downarrow$ \\
\midrule
Push-T & 5  & \textbf{0.017544} & 0.050309 & 0.094625 & 0.138354 \\
Push-T & 10 & 0.018656 & \textbf{0.048525} & 0.094867 & \textbf{0.133276} \\
Push-T & 15 & 0.021414 & 0.053420 & \textbf{0.093473} & 0.143585 \\
\midrule
Reacher & 5  & 0.000739 & 0.000926 & 0.001113 & 0.001296 \\
Reacher & 10 & \textbf{0.000735} & \textbf{0.000900} & \textbf{0.001075} & \textbf{0.001261} \\
Reacher & 15 & 0.000818 & 0.000959 & 0.001144 & 0.001345 \\
\bottomrule
\end{tabular}
\end{table}
$K=10$ is best on all reported Reacher horizons and gives the lowest Push-T error at $h=10$ and $h=20$, so we use it as the default span.


\subsection{CEM Score Controls and Tail Ablation}
\label{app:cem_score_control}

Table~\ref{tab:same_cem} reports end-to-end CEM performance under each model's frozen scoring rule. The controls below complement it with goal-only scoring and a PP-LeWM tail ablation.

\paragraph{CEM candidate scores.}
For candidate $m$, the ordinary terminal-goal score and Fast-LeWM self-consistency (SC) score~\citep{gao2026fastlewm} are
\begin{equation}
\begin{aligned}
J_{\mathrm{goal}}^{(m)}
&=\left\|\widehat{\vz}_{t+H}^{(m)}-\vz_g\right\|_2^2,\\
J_{\mathrm{Fast\text{-}SC}}^{(m)}
&=J_{\mathrm{goal}}^{(m)}+\beta\left\|\widehat{\vz}_{t+H}^{(m)}-\widetilde{\vz}_{t+H}^{(m)}\right\|_2^2.
\end{aligned}
\label{eq:cem_score_control}
\end{equation}
Here $\widetilde{\vz}_{t+H}^{(m)}$ is Fast-LeWM's re-anchored terminal estimate, while PP-LeWM uses the tail objective $J_\rho$ defined once in Eq.~(\ref{eq:coto_tail}). Setting $\beta=0$ recovers goal-only CEM, and $\rho=0$ gives the same terminal-goal ranking for PP-LeWM.

\paragraph{Controlled open-loop comparison.}
Table~\ref{tab:cem_score_control_all} combines the CEM score control with the PP-LeWM tail sweep and reports simulator Anytime and Terminal success.
\begin{table}[H]
\caption{Open-loop CEM score control and PP-LeWM tail ablation ($H{=}10$). Entries report Anytime and Terminal simulator success (\%). Fast-LeWM Goal removes SC, while PP-LeWM $\rho=0$ is ordinary terminal-goal CEM. Each configuration uses 100 episodes and three planner seeds.}
\label{tab:cem_score_control_all}
\centering
\resizebox{\textwidth}{!}{%
\begin{tabular}{@{}llcccccccc@{}}
\toprule
& & \multicolumn{2}{c}{Push-T} & \multicolumn{2}{c}{Reacher} & \multicolumn{2}{c}{Two-Room} & \multicolumn{2}{c}{OGBench-Cube} \\
\cmidrule(lr){3-4}\cmidrule(lr){5-6}\cmidrule(lr){7-8}\cmidrule(l){9-10}
Model & CEM score & Anytime$\uparrow$ & Terminal$\uparrow$ & Anytime$\uparrow$ & Terminal$\uparrow$ & Anytime$\uparrow$ & Terminal$\uparrow$ & Anytime$\uparrow$ & Terminal$\uparrow$ \\
\midrule
LeWM & Goal & 35.33 & 27.67 & 76.67 & 64.33 & 49.67 & 49.67 & 35.00 & 28.67 \\
Fast-LeWM & SC & 35.33 & 28.67 & 51.33 & 35.00 & 44.67 & 44.33 & 38.00 & 35.33 \\
Fast-LeWM & Goal & 30.00 & 22.33 & 51.67 & 34.33 & 50.67 & 50.67 & 36.33 & 30.00 \\
\midrule
PP-LeWM & $J_{0}$ (Goal) & 37.67 & \textbf{29.33} & \textbf{97.00} & \textbf{96.67} & 54.67 & 54.33 & 38.67 & 32.33 \\
PP-LeWM & $J_{0.1}$ & 35.67 & 25.00 & \textbf{97.00} & \textbf{96.67} & \textbf{55.33} & \textbf{55.00} & \textbf{45.67} & \textbf{39.33} \\
PP-LeWM & $J_{0.25}$ & 38.67 & 27.33 & 96.67 & 95.00 & 55.00 & \textbf{55.00} & 41.67 & 37.00 \\
PP-LeWM & $J_{0.5}$ & \textbf{41.33} & 26.67 & 96.33 & 93.67 & 52.33 & 52.33 & 43.67 & 38.33 \\
\bottomrule
\end{tabular}%
}
\end{table}

Under ordinary goal-only CEM ($J_0$), PP-LeWM already outperforms the goal-only LeWM and Fast-LeWM baselines on all four tasks, so the main planning gain does not depend on tail weighting.
Tail weighting shifts the operating point by task: it raises success on OGBench-Cube and Two-Room and increases Push-T Anytime success, whereas Reacher already saturates under $J_0$. Fast-LeWM's SC term is similarly task dependent.

\end{document}